\documentclass[conference]{IEEEtran}
\IEEEoverridecommandlockouts
\usepackage{cite}
\usepackage{amsmath,amssymb,amsfonts}
\usepackage{listings}
\usepackage{algorithmic}
\usepackage{graphicx}
\usepackage{textcomp}
\usepackage{url}
\usepackage{hyperref}
\usepackage{algorithm2e}
\usepackage{verbatim}
\usepackage{adjustbox}
\usepackage{xcolor}
\usepackage{fontawesome}
\graphicspath{ {./images/} }
\def\BibTeX{{\rm B\kern-.05em{\sc i\kern-.025em b}\kern-.08em
    T\kern-.1667em\lower.7ex\hbox{E}\kern-.125emX}}
    
    \definecolor{lightGray}{gray}{0.9}
\definecolor{Sepia}{rgb}{0.44, 0.26, 0.08}
\definecolor{Purple}{rgb}{0.7, 0.62, 0.71}
\definecolor{MidnightBlue}{rgb}{0.1, 0.1, 0.44}
\definecolor{LightGray}{rgb}{0.97,0.97,0.97}
\usepackage{xcolor}

\definecolor{codegreen}{rgb}{0,0.6,0}
\definecolor{codegray}{rgb}{0.5,0.5,0.5}
\definecolor{codepurple}{rgb}{0.58,0,0.82}
\definecolor{backcolour}{rgb}{0.95,0.95,0.92}

\lstdefinelanguage{SPARQL}{
  basicstyle=\small\ttfamily,
  backgroundcolor=\color{LightGray},
  columns=fullflexible,
  breaklines=true,
  sensitive=true,
  frame=bt,
  aboveskip=1em,
  belowskip=1em,
  xleftmargin=.5em,
  xrightmargin=.5em,
  framexleftmargin=.5em,
  framextopmargin=.5em,
  framexbottommargin=.5em,
  framexrightmargin=.5em,
  tabsize = 2,
  showstringspaces=false,
  morecomment=[l][\color{gray}]{\#},       
  morecomment=[n][\color{blue}]{<http}{>}, 
  morestring=[b][\color{OliveGreen}]{\"},  
  keywordsprefix=?,
  classoffset=0,
  keywordstyle=\color{Sepia},
  morekeywords={},
  classoffset=1,
  keywordstyle=\color{Purple},
  morekeywords={rdf,rdfs,owl,xsd,purl},
  classoffset=2,
  keywordstyle=\color{MidnightBlue},
  morekeywords={
    SELECT,CONSTRUCT,DESCRIBE,ASK,WHERE,FROM,NAMED,PREFIX,BASE,OPTIONAL, 
    FILTER,GRAPH, LIMIT,OFFSET,SERVICE,UNION,EXISTS,NOT,BINDINGS,MINUS,a
  }
}

\lstdefinelanguage{PDDL}
{
basicstyle=\small\ttfamily,
  backgroundcolor=\color{LightGray},
  columns=fullflexible,
  breaklines=false,
  frame=bt,
  aboveskip=1em,
  belowskip=1em,
  xleftmargin=.5em,
  xrightmargin=.5em,
  framexleftmargin=.5em,
  framextopmargin=.5em,
  framexbottommargin=.5em,
  framexrightmargin=.5em,
  tabsize = 2,
  showstringspaces=false,
  morecomment=[l][\color{gray}]{\#},       
  morecomment=[n][\color{blue}]{<http}{>}, 
  morestring=[b][\color{OliveGreen}]{\"},  
  sensitive=false,    
  keywordsprefix=:,
  morecomment=[l]{;}, 
  alsoletter={:,-},   
  morekeywords={
    define,domain,problem,not,and,or,when,forall,exists,either,
    :domain,:requirements,:types,:objects,:constants,
    :predicates,:action,:parameters,:precondition,:effect,
    :fluents,:primary-effect,:side-effect,:init,:goal,
    :strips,:adl,:equality,:typing,:conditional-effects,
    :negative-preconditions,:disjunctive-preconditions,
    :existential-preconditions,:universal-preconditions,:quantified-preconditions,
    :functions,assign,increase,decrease,scale-up,scale-down,
    :metric,minimize,maximize,
    :durative-actions,:duration-inequalities,:continuous-effects,
    :durative-action,:duration,:condition
  }
}

\begin{document}

\newcommand{\nb}[2]{
    \fbox{\bfseries\sffamily\scriptsize#1}
    {\sf\small\textcolor{red}{\textit{#2}}}
}
\newcommand\ag[1]{\nb{AG}{#1}}
\newcommand\da[1]{\nb{DH}{#1}}
 
\newtheorem{experiment}{Experiment}
\newtheorem{example}{Example}

\title{Handling abort commands for household kitchen robots}

\author{}

\author{\IEEEauthorblockN{Darius Has$^1$, Adrian Groza$^1$ and Mihai Pomarlan$^2$}
\IEEEauthorblockA{\textit{$^1$Department of Computer Science. Technical University of Cluj-Napoca}\\
\textit{$^2$Department of Applied Linguistics, University of Bremen, Germany}\\
}}

\maketitle

\begin{abstract}
We propose a solution for handling abort commands given to robots. 
The solution is exemplified with a running scenario with household kitchen robots. The robot uses planning to find sequences of actions that must be performed in order to gracefully cancel a previously received command.
The Planning Domain Definition Language (PDDL) is used to write a domain to model kitchen activities and behaviours, and this domain is enriched with knowledge from online ontologies and knowledge graphs, like DBPedia. 
We discuss the results obtained in different scenarios. 
\end{abstract}


\section{Introduction}

The task of aborting commands is essentially a problem of planning, or rather replanning whose core value comes when a robotic system is able to autonomously infer a fallback plan without a human in the loop.
A robot enhanced with capabilities of handling abort commands will able to reconfigure and replan its actions so that it can leave its environment in a clean state represents a step towards a more robust solution, given the fact that in the world of robotics malfunctions and unresponsiveness are risks that can be mitigated by having a fallback mechanism.

We developed here an \textit{Abort Task} module, as  
an enhancement to an existing robot simulator solution (i.e. AbeSim~\cite{}) and is designed to handle cancel or abort commands. Handling such commands is essential in improving the overall safety, robustness and reliability of the robots.
The proposed system is exemplified with a running scenario with household kitchen robots. 

Kitchen robots are able to execute simple tasks such as fetching kitchen vessels and utensils, cutting vegetables or even preparing a recipe. Ideally, they would also be able to clean after themselves or be able to recover and replan their actions once they receive a cancel or abort command so that the kitchen environment is left in a clean or safe state.


The paper is structured into six sections, starting with related work towards the task of cancelling, moving forward to the technical instrumentation section where the main technical apparatus will be showcased. Afterwards, the system's architecture and additional modules will be discussed, clarifying that the solution is an addition to an existing and functional robot simulator software able to perform tasks such as fetching cutlery and vessels, cutting or peeling vegetables or preparing a recipe. Special sections will be dedicated to running experiments and discussing results, as well as to providing an overview of the solution proposed along with an objective conclusion incorporating shortcomings and future improvements that may be taken into account. 

\section{Related Work}
 We briefly introduce related research on semantic knowledge as a means to increase the knowledge base of robotic systems assigned to perform and solve planning based problems. Afterwards, a brief explanation along with references will be provided towards using PDDL for solving planning problems, together with motivation for the employment of derived predicates by showcasing their benefits.

\subsection{Handling abort commands}
 The key challenges of handling cancel commands have been acknowledged by Haarland et al.~\cite{harland2017aborting}. These challenges include: the complex relationship between a goal and its subgoals, highlighting a need for a recursive approach for them, while also taking into account the plans in progress, i.e. the actual state of the world. 
 Handling such scenarios is increasingly important as cancel or abort commands can come at anytime, so the planning system needs to be reactive. 
 A clear differentiation does exist between: (i) dropping a goal or a task and (ii) aborting it, with the latter being characterized as having a plan on how to handle the command, while dropping meaning that no "clean-up" procedure is required. However, Haarland et al. have approached these to tasks as one, following the observation that  dropping a task means aborting it without having a sequence of steps to take. 

 Mora et al. have proposed a solution based on the Belief-Desire-Intention (BDI)\cite{mora1999bdi} agent architecture. 
 The BDI system includes abstract methods for handling abort commands, similar to a fallback mechanism. 
 One difference with our appriach is that goals 
 are not explicitly programmed with having abstract methods on how to recover from a current state, but rather the world is expressed by means of actions that can be taken, goals to be achieved and objects to interact with. 
 This means that the responsibility of replanning is considered as a search function to be solved by a planner based on, at most times, a heuristic or greedy search algorithm. 
 The function takes as inputs the current state, desired state and the capabilities of the robot and outputs a sequence of steps to be taken, meaning that handling of such abort commands is not explicitly programmed.

Handling an abort command for a robot can be seen as a change of context, change of goal for a robot, thinking about the fact that before receiving an external abort command, the robot's goal was to finish the in-progress action, while upon receiving the cancel command its focus should be on "repairing" the world state, so that it assures stability. 
This idea is explored by Fox et al.~\cite{fox2006plan}, where the concept of \textit{plan stability} is considered central to a planning system. 
Fox et al. have argued that no matter the cause of divergence between the previous goal and the one that arose, it is clear that the old plan must be replaced by a new one. 
They go on and present a differentiation between plan repairing as a strategy and replanning, further empirically concluding that the adaptation of the existing plan to a new context is more efficient than simply replanning from scratch. 
The paper introduces this contrast between plan repair and replanning in the context of the stability of a plan, which is considered a metric for showcasing adaptive capabilities of a robot. 
The empiric approach successfully presents the fact that high quality plans can be achieved in shorter time using the repair approach, while also maintaining the stability of the system when looking at the original plan, providing a clear indicator that the actual context or world state is essential, even when a shift in goals appears. This idea stands as a base to the approach presented in the later chapters, where the plan adapts and is reconstructed when the abort command is received. 

\subsection{Semantic representation in planning systems}
Semantic representation is as an aid for the planning system. 
In this line, Bernardo et al.~\cite{bernardo2021planning} describes  
how incorporating semantic knowledge into robotic systems can greatly enhance their capabilities. 
For example, by adding semantic knowledge to a robot's representation of its environment, it is able to make more informed decisions and perform tasks more efficiently. This goes along with the idea that incorporating ontologies with a planning system can provide more context, an idea that is further explored in the approach taken for the cancel task presented in the next sections.

Following the same line of augmenting procedural-like instructions such as those resulted from an AI planning system, but providing a vastly different approach, Pareti~\cite{pareti2018representation} presents an approach in which semantic knowledge can be used to extract information in a structured manner from a sequence of steps, i.e. a plan, in that way having a richer semantic representation with the scope of finding out what humans think and know by observing how they perform actions, showing how semantic ontologies can help increase the knowledge of a robotic system.

In the field of artificial intelligence, PDDL represents one of the most mature, abstract and widely considered solutions for solving problems that require the search of a sequence of actions to be taken. Yu-qian Jiang in \cite{jiang2019task} performs an experiment based comparison of two of the most widely used solutions for planning problems: planning domain definition language and answer set programming, concluding that, while each has its strengths, PDDL-based systems tend to behave better on problem with longer solutions, while ASP works better in an object crowded environment in which complex reasoning is required. Taking into consideration the task at hand, which is cancelling a kitchen related task and making sure that all the objects in the scene are left in a stable state, along with the fact that the objects in the scene are limited, PDDL represents a viable solution.

Several specifications of PDDL have been released, each with its set of additions and modifications to the already existing structure. One such addition is represented by axioms or derived predicates, which represent a special type of predicate that can be inferred by the planner from the current state, without being necessary to explicitly state it as an effect of an action or definition of the state. However, the planning community questioned the necessity of such axioms when talking about the ability of the planner to handle real-life domains and problems efficiently. Thiebaux et al.~\cite{thiebaux2005defense} have argued for the benefits of derived predicates 
and they also have probided 
an empirical presentation of the benefits in terms of decreasing the search space. 
and, as a result, promote efficiency, while also adding significant expressive power to PDDL specification. That is, the ability to more concisely express complex relationships.

\section{Technical instrumentation}
 This section starts by presenting the Abe Sim, the kitchen simulator used with the solution proposed. 
 Then, the PDDL is presenting 
 for expressing the world in which a planning problem resides, while showcasing the constraints and actions serving as guidelines for the robot. 
 Then, 
 an extension of PDDL will be explored - i.e. \textit{derived predicates} 
 and their benefits 
 in the context of handling abort commands.
Lastly, 
how ontologies can be used to augment data is proposed, as well as 
DBPedia can be used for the given task. 

\subsection{Abe simulator}
In a first solution using a robot simulator, Nevens et al. ~\cite{lrec-coling2024} have focused on 
natural language systems benchmark. The benchmark consists of 
a simulator in which a robot tries to follow the steps of cooking a meal following a recipe, in the way in which the NLU system understands them. If the steps are correctly inferred by the natural language system, they should be feasible for the robot and the outcome should be the one desired. 
Another possible application of a robot simulator is presented in~\cite{ringe2023taskbased}, in which AbeSim is used as a VR environment in which a human can interact with a robot, showing yet again the importance of the usage of a robot simulator, which on one hand provides a means of validating the correctness and completeness of a solution, while also mitigating the risks of having an actual robot constructed whose components may malfunction and with which the interaction would be harder to observe and troubleshoot.

\subsection{PDDL with derived predicates}
PDDL allows the abstraction of the domain of the world by 
formalising actions and predicates in order to express the state and its transitions. 

The primary components of PPDL are: (i) objects (entities that are manipulated by actions), 
(ii) predicates (the state of specific objects, under the closed world assumption); 
(iii) actions (for enabling transitions); 
(iv) initial state and (v) goal state.
Derived predicates 
represent a special types of predicates that can be used in order to add new predicates or states that need to be expressed, while also helping decrease the search space and, thus, as a consequence making the planner's work efficient, as previously stated in \cite{thiebaux2005defense}.

The main difference between a derived predicate and a standard predicate is the fact that the latter one needs to be explicitly expressed in order for it to hold true, while the derived one can be inferred to be true from the current state of the world. A very expressive example is the following: suppose we have a predicate \textbf{(at ?obj – locatable ?loc – locatable)} saying that an object is at a certain location. Now, let us say that we have an apple that is located in a bowl and the bowl is located in the fridge. Without derived predicates, in order for the predicate \textbf{(at ?apple ?fridge)} to be true, we need to express it explicitly. By having a derived predicate named \textbf{transitive-at} that would translate to the following explanation: an object is transitively at a location if it is at an intermediate container that is located at the target destination. Hence, specifying that the apple is in a bowl and the bowl is in a fridge, the planner can directly infer the fact that the apple is transitively-at the fridge.

This type of abstraction of the relationships or the states between the objects has the following advantages: they are more concise to write, they are more expressive while also being an aid in the representation complex relationships. What's more, they can be used in the goal definition of a PDDL problem, which transforms the problem of handling abort commands to a task of finding common sense rules for expressing the stability of the world using derived predicates that provide an indication of the clean states in which objects must reside and using those derived predicates to specify the goal of any problem in our bounded context, which is exactly the approach taken in this paper.
\subsection{Online ontologies. DBpedia}
Online ontologies contain structured knowledge extracted from various sources and are usually stored in RDF format,  which is considered to be the standard format for Semantic Web. Online ontologies are especially useful when we are talking about understanding, augmenting missing or incomplete data and enriching the capabilities of a robotic system.

DBpedia represents one of such existing ontologies. In order to query information, a special query language is used, called SPARQL. The main advantages of querying online ontologies such as DBpedia are:
   (i) structured information. Information that can be easily queried, interpreted and processed depending on the intent of its usage;
    (ii)  semantic relationships exist between entities. This can improve the knowledge and the understanding of the robotic system;
    (iii)  DBpedia is linked to other online ontologies, making it possible to have access to relationships that do not exist in DBpedia or are too abstractly expressed. 
DBpedia can aid when not enough information in the robot's knowledge base exists, especially in the case in which an action on a rather abstract object needs to be taken. By augmenting the existing data, context can be provided to the robot, further enhancing its knowledge along with its capabilities.
\section{System Architecture}
\begin{figure*}[!htb]
  \centering
  \includegraphics[width=\linewidth]{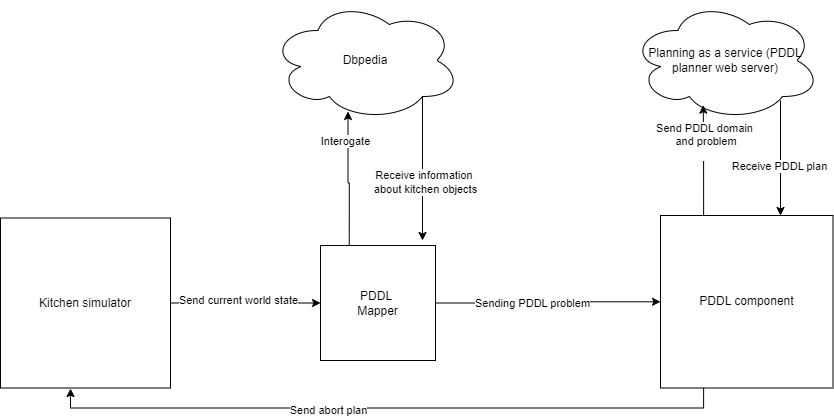} 
  \caption{System architecture}
  \label{fig:sysArh}
\end{figure*}

This section will present the overview of the system architecture together with the new modules added and explain their functionalities. One thing to mention from the beginning is the fact that the existing robot simulator was extended and integrated so that a PDDL based solution will be used when cancelling a command. The system architecture (Fig.~\ref{fig:sysArh}) presents the main module together with two new modules added for mapping the world extracted from the simulator to a PDDL problem and for actually containing the PDDL domain, receiving the PDDL problem computed dynamically and solving the actual problem. Each of these two added modules interact with external systems: the mapper interacts with DBpedia in order to augment data and help map an object from the scene to an object from the DBpedia domain and the PDDL module interacts with a server that contains different planners, each with different algorithm solutions whose puprose is to output a sequence of actions that leave the world in the desired state expressed in the goal definition.

\subsection{PDDL module}
The PDDL module is the one responsible for containing the domain of the kitchen and for handling the communication with the planner server so that the domain and problem data is sent to it and the sequence of actions is received and processed so that it is forwarded to the existing simulator module for handling commands.

The first thing worth mentioning is the domain that contains the types of objects present in the kitchen: \textbf{storage} ("clopenable" or "notclopenable", the difference between them being the fact that a "clopenable" storage can be opened or closed), \textbf{device}, \textbf{fridge} (a special kind of device), \textbf{vessels}, \textbf{utensils} and \textbf{perishable} and \textbf{non-perishable} food. All of these objects have to be of an abstract type named \textbf{locatable} in order for them to be marked as being able to be located in the scene at a certain location. 
The entire hierarchy of objects can be observed in Listing~\ref{lst:objects}.
\begin{lstlisting}[label = lst:objects,caption=Objects in the kitchen scenario,language=PDDL]
(:types
locatable - object
container utensil food disposable - locatable
vessel storage trash_can - container
clopenable notclopenablestorage - storage
device fridge clopenablestorage - clopenable
perishable nonperishable - food)
\end{lstlisting}

Another component of the domain is represented by actions that can be taken in the world. These actions are:
\begin{itemize}
    \item \textbf{Close}. Used for closing an opened storage.
    \item \textbf{Move}. It is used for moving a locatable item from one location to another.
    \item \textbf{Turn-off}. Action used for turning off devices.
    \item \textbf{Put}. A specific action used if the robot is holding something and needs to put it down in a container.
\end{itemize}

All these actions have been expressed as PDDL actions, which means that they contain as parameters objects that can appear in the world, as preconditions some predicates that state that the world is in a state in which the action can be performed and the effects stating the transition to the new state of the world for the objects affected by the action.

An example of an action implementation is described in the listing \ref{lst:action}, where we can see that in order to move an object the robot needs to be able to grasp the object and needs to be at the source location and needs to be graspable. The effects state that the object is no longer at the source location but rather is stored in a destination container.

\begin{lstlisting}[label = lst:action, caption=Declaration of the move action in PDDL,language=PDDL]
(:action move
        :parameters (?gr - locatable
                     ?src - container
                     ?dest - storage)
        :precondition (and
            (robot-can-grasp)
            (not (immobile ?gr))
            (at ?gr ?src))
        :effect (and
            (not (at ?gr ?src))
            (at ?gr ?dest)))
\end{lstlisting}

Last but not least, the core of this PDDL implementation is the usage of derived predicates. For each of the major categories stated above, there exists a derived predicate stating that the object of that certain type is in a safe state. For example, for a perishable food, it is considered that it is safely stored if it is located in the fridge. Similarly, each vessel or each utensil have safe derived predicates specifying common sense rules to be considered in order for the kitchen to be considered in a safe state. These derived predicates are also useful to specify a transitive at relationship, which can be used to infer that a food is at a fridge without explicitly having a declared predicate or an action's effect that states that. This is really useful and it is used in all other derived predicates, so that items do not have to be directly stored at the destination, but can also be located in a container that is at the destination. Listing \ref{lst:derived} presents some derived predicates present in the PDDL domain.

\begin{lstlisting}[label = lst:derived, caption=Derived predicates,language=PDDL]
(:derived (robot-can-grasp)
    (and (not (exists
                (?obj - locatable)
                (holding-left ?obj)))
         (not (exists
                (?obj - locatable)

(:derived
        (safe-perishable ?p - perishable)
        (exists
            (?fr - fridge)
            (transitive-at ?p ?fr)))
\end{lstlisting}

All these derived predicates will be used in the definition of the goal state of the problem, meaning that for any problem inferred from the world state by the mapper module, the goal will always be that all the kitchen objects are in a safe, stable state.
\subsection{Abe PPDL mapper module}
This module is responsible for taking the current world state of the abe and creating the initial state for the PDDL problem. This means that for each object, the mapper needs to infer its type and needs to infer the location of it so that it is specified in the PDDL problem. What's more, certain special characteristics need to be inferred also, such as: what objects the robot is holding in its hand, or whether a container can be moved or not. Furthermore, for devices or containers that can be opened, or turned off and on, the mapper needs to make sure that information about them being open, closed or on and off is inferred and mapped to the PDDL problem.

The mapper reads the world state from the simulator at the moment in which the cancelling request is sent and goes through each object and its particularities in order to make sure that it can extract all the information necessary. The type of the object is added to the PDDL problem and then characteristics of interest, such as location, movability, the fact that the object is closed or not or turned on or off are mapped to the problem using the available predicates.

There exists a direct correlation between the characteristics of an object in a simulator and their corresponding types in PDDL, which is used for this exact mapping. If, however, no information from the abe world is useful, the mapper uses DBpedia as a fallback mechanism for accessing necessary information in order to map the object to its corresponding type. An example of this mapping is the following: in the abe simulator any object that has characteristics such as 'canBake' or 'canCut' is considered to be a utensil. Similarly, for other PDDL types there exists a direct mapping that can be used. 

There exists certain objects, usually meals, that do not contain any characteristics in PDDL apart from the fact that they can be grasped by the robot. Hence, custom SPARQL queries were written in order to interact with DBpedia and extract information about meals, in this case. 
Listing ~\ref{lst:sparql1} presents how we can extract this kind of information, stating that meals, which are a special type of perishable food are considered to be every food from DBpedia that contains ingredients. There is an union with vegetables and fruits so that a whole set of perishable foods is inferred from DBPedia. This is the procedure in which the data extracted from the Abe world is augmented when needed so, in order for the mapper to be able to infer the type of a certain object from the scene. 
\begin{lstlisting}[language=SPARQL, label = lst:sparql1, caption=Extracting perishable food from DBpedia]
SELECT DISTINCT ?thing 
WHERE {
    {?thing rdf:type dbo:Food .
    ?thing dbo:ingredient ?ingredient .}
    UNION {?thing dcterms:subject dbc:Edible_fruits .}
    UNION {?thing dcterms:subject dbc:Fruit_vegetables .}
    UNION {?thing dcterms:subject dbc:Root_vegetables .}}
\end{lstlisting}
Similarly, listing  ~\ref{lst:sparql2} presents the method in which a set of utensils is computed from DBpedia. It can clearly be seen that the subject relation is taken into account by querying for RDF tuples which contain a subject of Cooking vessels or cookware and bakeware, and extracting all these subjects as a set of utensils. 
\begin{lstlisting}[language=SPARQL, label = lst:sparql2, caption=Extracting utensils from DBpediaa]
SELECT DISTINCT ?thing 
WHERE {
    {?thing dcterms:subject dbc:Cookware_and_bakeware .}
    UNION
    {?thing dcterms:subject dbc:Cooking_vessels .}}
\end{lstlisting}

\section{Running Experiments} 
We considered two scenarios involving command cancellations: (1) Moving a bowl to the kitchen counter, and (2) Cutting an onion.
For both experiments, there will be two parts that need to be taken into account and analyzed. These are the degree to which the world has been correctly mapped to a PDDL problem and the actual plan that the planning service has inferred using the domain provided with all the knowledge necessary and the problem that has been mapped from the actual state of the world of the simulator. What is more, the scene will be consistent between the two running scenarios, meaning that objects and their placement will differ only slightly. The particularities of the scene are: vessels and utensils, such as a knife will be present at the kitchen cabinet, a fridge, a kitchen counter and a kitchen cabinet exist as storage, an oven and a couple immovable parts exists to conclude the structure of the scene, such as the floor and the walls. 

In the experiments, only the characteristics of each scenario will be analysed and a discussion will be made after presenting the resulted plans.

Before specifically analysing each scenario, one thing is to be noted, and that is the planning configuration used to output a plan. The planning system used is based on the Fast-Downward~\cite{helmert2006fast} framework, more notably the LAMA-first planning system. This planning system aims to heuristically solve a search problem using a greedy approach, based on best first search. LAMA-first disregards the refinements performed in a regular LAMA \cite{richter2011lama} model and hence, it does not take into account the cost of finding the solution. While the optimal plan is not guaranteed to be achieved because of the greedy best-first search approach, this planner configuration (Fast-Downward with LAMA-first) represents the only one in the planner service used that supports PDDL 2.1 specification and hence, derived predicates. 
This was the main factor in favor of choosing this planner and as a direct result, because of the shortage of planning systems supporting derived predicates. 
A analysis regarding the impact on the performance of the system can be done by employing different heuristics available in the Fast-Downward syste.. 

\begin{figure*}
  \centering
  \includegraphics[width=\linewidth]{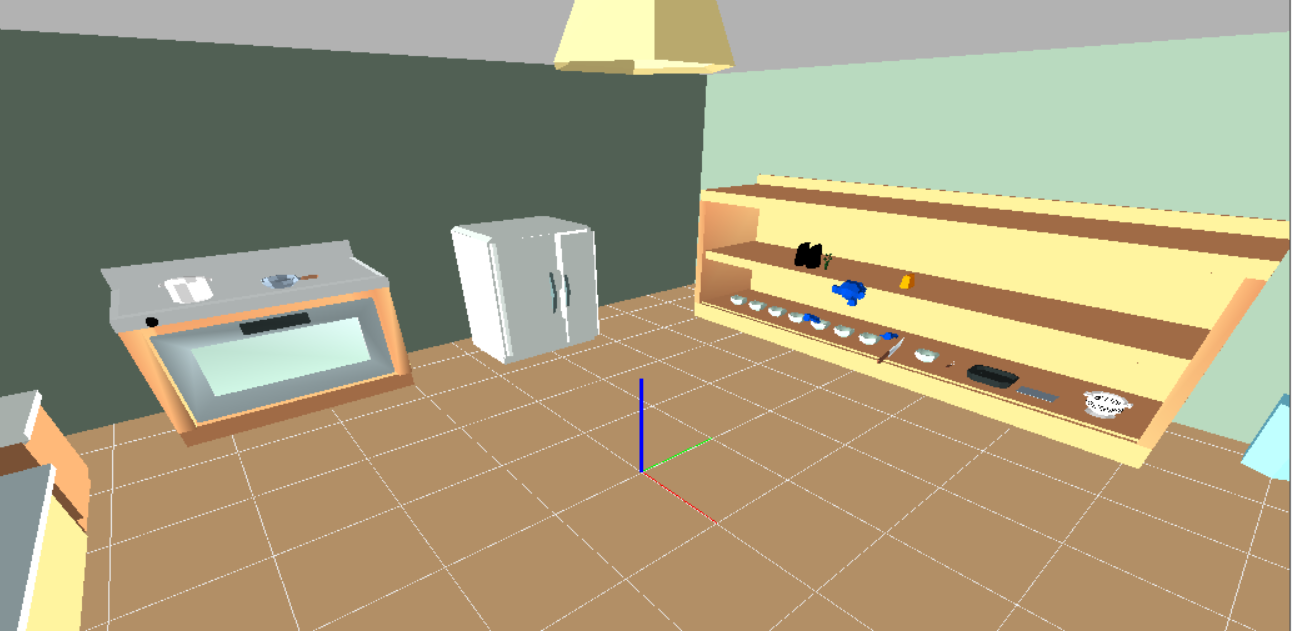} 
  \caption{Fetching a knife in the Abe Sim}
  \label{fig:abe}
\end{figure*}

\subsection{Running scenario 1. Moving a bowl to the kitchen counter}
The robot is instructed to fetch a bowl. It goes and grab it and when it is on its way to the desired location of the bowl, an abort instruction is received.
At the moment of receiving the cancel command, the world has been translated into a PDDL problem (Listing~\ref{lst:scenario1}), with relevant information about the bowl being shown as: the medium bowl was inferred to be a vessel and it was inferred to be held in the right hand by the robot, indicated by the holding-right predicate.

The same listing shows the resulted plan, which consists of only one step: to put the medium bowl to the kitchen counter.
\begin{lstlisting}[label=lst:scenario1,caption=PDDL problem results 
in the first running scenario,language=PDDL]
medium_bowl3 - vessel
(holding-right medium_bowl3)

Plan resulted:
('put', ['mediumBowl3', 'kitchenCounter'])
\end{lstlisting}

\subsection{Running scenario 2. Cutting an onion}
This scenario is more complex. There is only one command that precedes the cancellation command, which is exactly that of cutting the onion. The robot knows the location of a knife, the location of the onion and moves towards the location of the knife. After the knife is picked-up and the robot intends to go to the kitchen counter, where the onion resides, it is prompted yet again with the cancel command. 

Listing~\ref{lst:scenario2} presents, as with the previous scenario, relevant information about the important objects in the scene, i.e. objects relevant for this scenario, as long as the plan provided. As seen, the mapper correctly infers the types of both the onion and the knife, considering them to be a perishable food and an utensil, which is in line to the classification made by humans using common sense. The interesting parts comes when looking at the resulting plan, which consists of two steps. The first step is similar to the one taken in the first scenario, which is to put the knife, held in the right hand, on the kitchen counter. The second step shows the intent to move the onion to the fridge, which goes in line with the knowledge base constructed using derived predicates, that clearly specifies that all perishable foods need to reside in the fridge in order for them to be stored safely.

\begin{lstlisting}[label = lst:scenario2, caption=PDDL problem results
in the second running scenario,language=PDDL]
cooking_knife - utensil
onion - perishable

(holding-right cooking_knife)
(at onion kitchen_counter)

Planner finished in 6.552 seconds.
Plan resulted:
('put', ['cookingKnife', 'kitchenCounter']),
('move', ['onion', 'kitchenCounter', 'fridge'])
\end{lstlisting}

\textit{Code and data availability:} the code is available at \textit{Blind}.

\subsection{Results discussion}
While the scenarios indicate that the fallback mechanism is indeed present and the PDDL module is able to translate the world correctly into a PDDL problem and provide a sequence of actions steps to be taken to leave the kitchen and its object in a clear state, it is also worth stating the fact that there is no single and hence, no optimal plan provided. In the first step of both scenarios, an object is fetched from the kitchen cabinet and after it is grabbed by the robot, the cancel command is received and in both, the first action step is to put the object at the kitchen counter. This represents an arbitrary decision taken by the robot, because in PDDL there is no difference between the two storage spaces. This represents a first shortcoming of the solution mentioned above, because there could be cases in which certain storage spaces to be more suitable to store certain objects. What is more, there is no info about the storage's occupation and hence, there could be cases in which the planner indicates that an object should be stored at a specific location, without having the context about the storage's fullness.

Another observation is the fact that right now, the system is limited to having certain unique objects, such as the fridge. It goes without saying that adding more objects to the scene will increase the search space and will make the search algorithm more costly. What's more, because the system has been tested with only one planner due to the need of support for derived predicates, there is no metric used to measure the quality of the result. One future improvement would be to search for several planners capable of handling derived predicates and using them in order to compare and measure the cost for each of the solutions provided in order to select the best, most cost effective solution. 

\section{Conclusion}
We proposed here a solution for the handling abort commands by robots.
The robot is able to translate the world state into a PDDL problem, to infer a plan and to execute it in order to leave the objects in the kitchen in a clean, stable state. 
The derived predicates extension of PDDL comes with the main benefit of being able to abstractly, but concisely express commonsense rules for different types of objects to be taken into account and be used in the goal specification of the closed world, ensuring that the system is provided with a knowledge base of the relevant objects and their desired state. 
Ontologies are used to augment missing data between the state of the robot or the simulator and the necessary information. 

The interesting researcher can improve the current implementation by: (i) adding information about the state of the storage so that no steps that cannot be taken are inferred (e.g. moving an object into a full storage(l (ii)  
using derived predicates that take into account cost 
(iii) extending the hierarchy of objects in the PDDL domain, a
or having more commonsense knowledge being expressed in the form of derived predicates. These represent steps towards enriching the knowledge base of the PDDL system.

\bibliographystyle{IEEEtran}
\bibliography{main}
\end{document}